# Advancing Roadway Sign Detection with YOLO Models and Transfer Learning

Selvia Nafaa
*Computers and Systems Department*
*Minia University*
Minia, Egypt

Karim Ashour
*Computers and Systems Department*
*Minia University*
Minia, Egypt

Rana Mohamed
*Computers and Systems Department*
*Minia University*
Minia, Egypt

Hafsa Essam
*Computers and Systems Department*
*Minia University*
Minia, Egypt

Doaa Emad
*Computers and Systems Department*
*Minia University*
Minia, Egypt

Mohammed Elhenawy
*CARRS-Q*
*Queensland University of Technology*
Brisbane, Australia
mohammed.elhenawy@qut.edu.au

Huthaifa I. Ashqar
*Civil Engineering Department*
*Arab American University*
Jenin, Palestine
huthaifa.ashqar@aaup.edu, 0000-0002-6835-8338

Abdallah A. Hassan
*Computers and Systems Department*
*Minia University*
Minia, Egypt
abdallah@mu.edu.eg

Taqwa I. Alhadidi
*Civil Engineering Department*
*Al-Ahliyya Amman University*
Amman, Jordan
t.alhadidi@ammanu.edu.jo

*Abstract*—Roadway signs detection and recognition is an essential element in the Advanced Driving Assistant Systems (ADAS). Several artificial intelligence methods have been used widely among of them YOLOv5 and YOLOv8. In this paper, we used a modified YOLOv5 and YOLOv8 to detect and classify different roadway signs under different illumination conditions. Experimental results indicated that for the YOLOv8 model, varying the number of epochs and batch size yields consistent MAP50 scores, ranging from 94.6% to 97.1% on the testing set. The YOLOv5 model demonstrates competitive performance, with MAP50 scores ranging from 92.4% to 96.9%. These results suggest that both models perform well across different training setups, with YOLOv8 generally achieving slightly higher MAP50 scores. These findings suggest that both models can perform well under different training setups, offering valuable insights for practitioners seeking reliable and adaptable solutions in object detection applications.

## I. Introduction

In intelligent transport and automotive technology, detecting and identifying roadside signs is fundamental. Nowadays, progress has been substantial in this realm, due to the advancement in the power of computer vision. Traffic management and road safety are both helped by that idleness: computer vision makes it possible for automatic systems to recognize and understand roadway signs [1]. Within this new realm, the underpinning technology is to utilize digital images produced from vehicle-mounted or roadside cameras to in real-time detect traffic-signs and classify them simultaneously [2]. These systems use advanced algorithms such as image processing techniques and machine learning models to accurately detect various sign types, shapes, and symbols in different environmental and lighting conditions [3].

A key part of the process is the application of convolutional neural networks (CNNs), which have been shown to effectively recognize and classify images [4]. This network is not much troubled by the visual complexity of roadway signs, and is able to divide between regulatory, warning, and information signs and the specific information they convey [5].

To develop fully autonomous vehicles in the transportation sector--feeding off machine vision technology as an extension might be helpful for developing humans. However, protection measures are still necessary. Error rates at current low levels could lead to serious accidents or death without them. In this respect, computer vision technology is helpful in identifying and interpreting signs as they become more sophisticated. This is especially important when communication is concerned since public and private transportation are not environmentally safe. The integration of computer vision in roadway sign detection is not only essential for the development of autonomous vehicles but also holds great opportunity to serve human drivers [6], [7]. This technology enables more timely and effective responses to road signs, reducing the likelihood of accidents and easing traffic flow out of town. In this work, we used YOLOv5 and YOLOv8 to detect and classify roadway signs under different illumination conditions.

## II. Literature Review

An array of research has set its sights on using methods like deep learning and convolutional neural networks (CNN) to detect and recognize street signs. Another study [8] provided an in-depth look at traffic signal identification, monitoring, sorting out problems, including the latest trends and challenges associated therewith. The authors also note that machine- and vision-based systems have made advancements and admit that there are still some areas we need to improve. In [9], [10], they submitted a study on the 3D localization of traffic signs with 2D detect-classify from overlapping images, and this method is to show how a synthesis in modern technology can increase accuracy reliability of sign recognition systems. Furthermore, another study [11] presents a method in machine learning for reading damaged arrow-road markings using CNN technology. In [9], the study illustrated 3D localization of traffic signs in 2D detection and classification model from overlapping images. This approach can integrate advanced technology to enable higher precision and a safer working experience in roadway sign detection. Other authors such as[12], [13] have examined the use of CNN models to classify traffic signs and make them visible. They show the profound importance of deep learning technologies for achieving high accuracy in named entity recognition. The potential of deep learning technologies through CNN methods has also been indicated in these studies. Over the last few years, it's been shown that deep learning can be applied to traffic sign detection [14]. Their research submits a number of encouraging conclusions. They also introduced a real-time traffic sign detection algorithm for high-performance. On the way, it demonstrates that road sign automatic recognition can be done with the help of feature extraction and computer vision technology. These references bring together deep learning, CNN, and machine vision systems in the field of highway sign classification and identification. Looking at these methods, the outlook for future improvements in accuracy is excellent.

In [15], the authors introduced an updated YOLOv5 method for using the enhanced YOLOv5 iteration to recognize traffic signs. The paper emphasizes the importance of YOLO-based models in quickly identifying traffic signs in context. This resource serves as an invaluable piece of information for applying YOLOv5 to recognize certain objects in road networks. However, in [16], authors introduced SF-YOLOv5, a technique using a compact method to detect small objects. The method employs an upgraded feature integration model. This study reflects the development of YOLOv5 in detecting small objects well, especially in visually difficult conditions, for the job of accurately identifying roadside signs. Nonetheless, another study [17] developed and designed a deep learning algorithm that is capable of identifying and classifying traffic signs in a loop. In addition, the paper experimentally shows that AI technologies can be used to create power platforms. Moreover, papers have drawn attention to the capability of YOLO models for actually making road signage detection nice things that work out in real life situations.

### III. METHODOLOGY

In this section, the proposed research methodology is presented. The aim of this study is using computer vision to create a digital map that shows the roadway signs, information about road signs, and other features of the road environment. Assuming that images and locations for signs are collected using emerging technology tools (i.e., connected vehicles with LIDAR system). Then, the collected data can be modeled using the proposed methodology in classifying the different roadway assets. The data used are open-access data downloaded from Kaggle. In the classification of road signs, we used the tiny LISA dataset of nine classes and total images of 900. This dataset is a subset from the German Traffic Sign Recognition Benchmark (GTSRB) dataset which contains 43 classes from different road signs. The images cover different light, weather, and other driving conditions. Both standalone frames and videos are part of the LISA dataset. Although not all frames have been extracted for annotation, every frame that has been marked may be linked back to the original video. As a result, the annotations can be used to validate systems using tracking. The detailed preprocessed methodology is shown in Fig. 1.

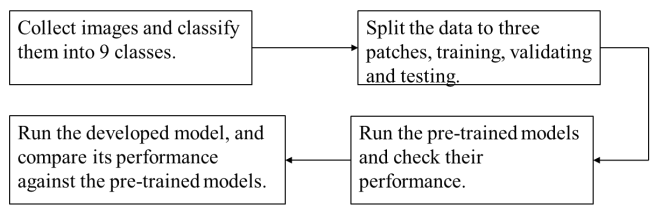

Fig. 1. Chartflow for the preprossecing steps.

Fig. 2 shows the imbalance of the original dataset classes. We augmented our dataset in order to balance it and from data preparation work we also resized the dataset images to fit the various classification models. In image recognition, there are two main approaches to transfer learning: fine-tuning and feature extraction [18], [19]. The weights of a pre-trained model are further trained on a new dataset of images during fine-tuning. This allows the model to adapt to the specific features of the new dataset without having to learn from scratch. Feature extraction, on the other hand, employs the pre-trained model as a fixed feature extractor by removing a few final layers and training a new model on the extracted output. This method is especially useful for smaller datasets with limited training data. Furthermore, feature extraction allows for faster training times because the pre-trained model has already learned a significant amount of information from the original dataset. For the first approach, we used YOLOv5 and YOLOv8. However, A feature extraction technique was used in this study to extract image embeddings from the new dataset. Pre-trained models, primarily Inception, ImageNet, and AlexNet, were used for this purpose. The image embedding output is typically a high-dimensional vector containing numerical feature values that can be used for a variety of tasks.

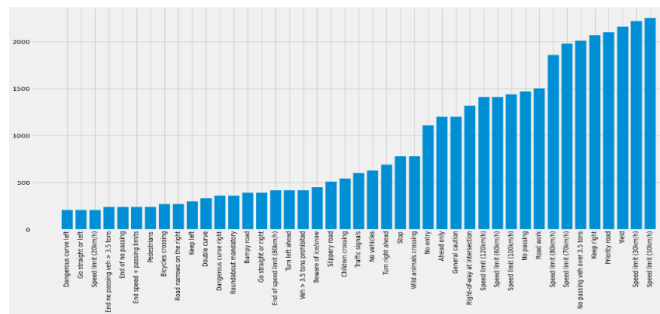

Fig. 2. Classes of GTSRB dataset and number of images in each class.

### IV. ANALYSIS AND RESULTS

In this section, using YOLOv5 and YOLOv8, the different evaluation metrics were reported for the three dataset splits including training, validation, and testing under different epochs size and batch. Results indicated that the model performance is robust to the batch and epochs sizes. We used Mean Average Precision at the top 50 (MAP50), which is a performance metric commonly used in information retrieval, particularly in the evaluation of object detection and image retrieval algorithms. MAP50 evaluates the precision of a model's predictions when considering the top 50 ranked items. The precision at each rank is calculated by dividing the number of relevant items retrieved by the total number of retrieved items up to that rank. Average Precision (AP) is then computed by averaging the precision values at each relevant item rank. The Mean Average Precision (MAP) is obtained by averaging the AP scores across all queries or instances.

These results are shown in Table 1. The confusion matrix for YOLOv5 and YOLOv8 is shown in Fig. 3. The results showcase the performance of two object detection models, YOLOv8 and YOLOv5, across different training configurations and datasets. For the YOLOv8 model, varying the number of epochs and batch size yields consistent MAP50 scores, ranging from 94.6% to 97.1% on the testing set. The YOLOv5 model demonstrates competitive performance, with MAP50 scores ranging from 92.4% to 96.9%. These results suggest that both models perform well across different training setups, with YOLOv8 generally achieving slightly higher MAP50 scores.

Table 1. Evaluation metrics for roadway signs detection using YOLOv5 and YOLOv8.

| Model | Epoch | Batch | MAP50 | | |
|---|---|---|---|---|---|
| | | | Training | Validation | Testing |
| YOLOv8 | 100 | 64 | 96.3 | 96.3 | 94.8 |
| YOLOv5 | 100 | 64 | 95.7 | 93.7 | 96.9 |
| YOLOv8 | 100 | 16 | 96.3 | 96.3 | 95 |
| YOLOv5 | 100 | 16 | 93.6 | 92.6 | 96.1 |
| YOLOv8 | 90 | 16 | 97.1 | 97.0 | 94.6 |
| YOLOv5 | 90 | 16 | 94.4 | 92.4 | 96.3 |

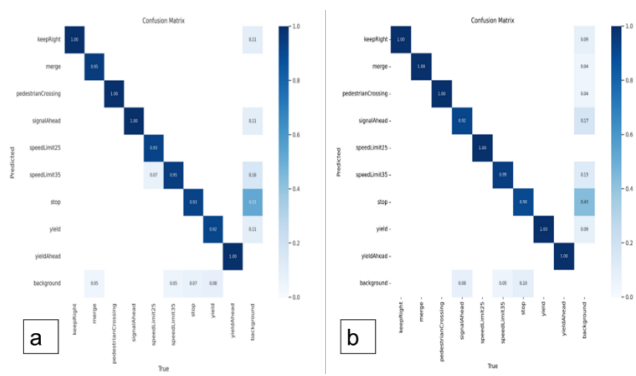

Fig. 3. Confusion Matrix at batch size 16 and epochs of 90 for (a) YOLOv8 and (b) YOLOv5.

This research has shown that sophisticated deep learning algorithms, in particular YOLOv5 and YOLOv8, have the potential to be used for the identification and categorization of traffic signs. Based on the findings, we found that the implementation of YOLOv5 and YOLOv8 has demonstrated high accuracy in roadway signs classification.

## V. CONCLUSION

A roadway asset identification method based on a deep convolutional neural network fusion model was implemented in this study. The model validation strategy was carried out by testing the performance of several models for detecting roadway signs, effectively improving its asset detection accuracy. In summary, the following conclusions can be drawn from this research. We implemented YOLOv5 and YOLOv8 in roadway signs detection and classification for roadway asset management.

The evaluation of YOLOv5 and YOLOv8, utilizing different epochs and batch sizes across three dataset splits (training, validation, and testing), reveals a robust performance of the models. The models demonstrated consistency in performance across various training configurations, highlighting their resilience to changes in batch and epochs sizes. The provided results underscore the effectiveness of both YOLOv8 and YOLOv5 in object detection tasks. Notably, YOLOv8 exhibited slightly higher MAP50 scores, ranging from 94.6% to 97.1%, on the testing set, while YOLOv5 demonstrated competitive performance with scores ranging from 92.4% to 96.9%. These findings suggest that both models can perform well under different training setups, offering valuable insights for practitioners seeking reliable and adaptable solutions in object detection applications. Additionally, considerations such as training time, resource utilization, and specific use case requirements should be factored in when determining the most suitable model for practical implementations.

REFERENCES

[1] S. Papagianni, C. Iliopoulou, K. Kepaptsoglou, and A. Stathopoulos, "Decision-Making Framework to Allocate Real-Time Passenger Information Signs at Bus Stops: Model Application in Athens, Greece," *Transportation Research Record Journal of the Transportation Research Board*, 2017, doi: 10.3141/2647-08.

[2] P. Arnoul, M. Viala, J. P. Guerin, and M. Mergy, "Traffic signs localisation for highways inventory from a video camera on board a moving collection van," in *Proceedings of conference on Intelligent Vehicles*, IEEE, 1996, pp. 141–146.

[3] R. C. Gonzalez and R. E. Woods, "Digital image processing, prentice hall," *Upper Saddle River, NJ*, 2008.

[4] Z. Zhu, D. Liang, S. Zhang, X. Huang, B. Li, and S. Hu, "Traffic-sign detection and classification in the wild," in *Proceedings of the IEEE conference on computer vision and pattern recognition*, 2016, pp. 2110–2118.

[5] C. Ai and Y. J. Tsai, "An automated sign retroreflectivity condition evaluation methodology using mobile LIDAR and computer vision," *Transp Res Part C Emerg Technol*, vol. 63, pp. 96–113, 2016.

[6] R. P. Loce, R. Bala, M. Trivedi, and J. Wiley, *Computer vision and imaging in intelligent transportation systems*. Wiley Online Library, 2017.

[7] T. Strain, R. Eddie Wilson, and R. Littleworth, "Computer vision for rapid updating of the highway asset inventory," *Transp Res Rec*, vol. 2674, no. 9, pp. 245–255, 2020.

[8] S. B. Wali, M. Abdullah, M. A. Hannan, A. Hussain, S. A. Samad, and M. Mansor, "Vision-Based Traffic Sign Detection and Recognition Systems: Current Trends and Challenges," *Sensors*, 2019, doi: 10.3390/s19092093.

[9] V. Balali, A. Ashouri Rad, and M. Golparvar-Fard, "Detection, classification, and mapping of US traffic signs using google street view images for roadway inventory management," *Visualization in Engineering*, vol. 3, pp. 1–18, 2015.

[10] M. Golparvar-Fard, V. Balali, and J. M. de la Garza, "Segmentation and recognition of highway assets using image-based 3D point clouds and semantic Texton forests," *Journal of Computing in Civil Engineering*, vol. 29, no. 1, p. 04014023, 2015.

[11] H. Vokhidov, H. G. Hong, J. K. Kang, T. M. Hoang, and K. R. Park, "Recognition of Damaged Arrow-Road Markings by Visible Light Camera Sensor Based on Convolutional Neural Network," *Sensors*, 2016, doi: 10.3390/s16122160.

[12] Y. Chew *et al.*, "Malaysia Traffic Signs Classification and Recognition Using CNN Method," 2022, doi: 10.3233/atde221120.

[13] W. Ding, W. Li, and D. Ma, "Design of Improved Road Sign Recognition System Based on Deep Learning," 2023, doi: 10.1117/12.2667330.

[14] J. Chu, C. Zhang, M. Yan, H. Zhang, and T. Ge, "TRD-YOLO: A Real-Time, High-Performance Small Traffic Sign Detection Algorithm," *Sensors*, 2023, doi: 10.3390/s23083871.

[15] B. Zheng, H. Lv, and H. Lu, "Traffic Sign Detection Algorithm Based on Improved YOLOv5," 2022, doi: 10.1117/12.2659655.

[16] H. Liu, F. Sun, and L. Deng, "SF-YOLOv5: A Lightweight Small Object Detection Algorithm Based on Improved Feature Fusion Mode," *Sensors*, 2022, doi: 10.3390/s22155817.

[17] N. Kargah-Ostadi, A. Waqar, and A. Hanif, "Automated Real-Time Roadway Asset Inventory Using Artificial Intelligence," *Transportation Research Record Journal of the Transportation Research Board*, 2020, doi: 10.1177/0361198120944926.

[18] H. I. Ashqar, M. Elhenawy, M. Masoud, A. Rakotonirainy, and H. A. Rakha, "Vulnerable road user detection using smartphone sensors and recurrence quantification analysis," in *2019 IEEE Intelligent Transportation Systems Conference (ITSC)*, IEEE, 2019, pp. 1054–1059.

[19] M. Elhenawy, H. I. Ashqar, M. Masoud, M. H. Almannaa, A. Rakotonirainy, and H. A. Rakha, "Deep transfer learning for vulnerable road users detection using smartphone sensors data," *Remote Sens (Basel)*, vol. 12, no. 21, p. 3508, 2020.

.